# WHO-Hand Hygiene Gesture Classification System

Rashmi Bakshi

*Abstract* The recent ongoing coronavirus pandemic has highlighted the importance of hand hygiene practices in our daily lives, with governments and health authorities around the world promoting good hand hygiene practices. More than 1 million cases of hospital-acquired infections are reported in Europe annually. Hand hygiene compliance may reduce the risk of cross- transmission thereby reducing the number of infections as well as health-care expenditures. In this paper, WHO hand hygiene gestures were recorded and analyzed with the construction of an aluminum frame, placed at the laboratory's sink. The hand hygiene gestures were recorded for thirty participants after conducting a training session about hand hygiene gestures demonstration. The video recordings were converted into image files and were organized in six different hand hygiene classes. The Resnet-50 framework was selected for the classification of multi-class hand hygiene stages. The model was trained with the first set of classes (Fingers Interlaced, P2PFingers Interlaced, and Rotational Rub) for 25 epochs. An accuracy of ~44% was achieved for the first set of experiments with loss score >1.5 in validation set. The training steps for the second set of classes (Rub hands palm to palm, Fingers Interlocked, Thumb Rub) were 50 epochs. An accuracy of ~72% was achieved for the second set with the loss score < 0.8 for the validation set. In this work, a preliminary analysis of robust hand hygiene dataset with transfer learning was carried out with a future aim of deploying a hand hygiene prediction system for healthcare workers in real-time.

## I. Introduction

Hospital Acquired Infections (HAIs) have a significant impact on quality of life and result in an increase in health care expenditure. According to the European Centre for Disease Prevention and Control (ECDC), 2.5 million cases of HAIs occur in European Un-ion and European Economic Area (EU/EAA) each year, corresponding to 2.5 million DALYs (Disability Adjusted Life Year) which is a measure of the number of years lost due to ill health, disability or an early death [1]. MRSA-Methicillin Resistant Staphylococcus Aureus is a common bacteria associated with the spread of HAIs [2].

One method to prevent the cross transmission of these microorganisms is the implementation of well-structured hand hygiene practices. The World Health Organization (WHO) has provided guidelines about hand washing procedures for health care workers [3]. Best hand hygiene practices have been proven to reduce the rate of MRSA infections in a health care setting [4].

One challenge in dynamic healthcare environments is to ensure compliance with these hand hygiene guidelines and to evaluate the quality of hand washing. This is often done through auditing involving human observation. The hand washing process, however, is well structured and has particular dynamic hand gestures associated with each hand washing stage.

The assessment of the process may therefore be suited to automation. Existing technology includes the use of electronic counters and RFID badges to measure the soap usage and location based reminder systems to alert the workers about washing hands [5, 6, 7]. These systems have shown to improve the frequency of hand washing but they do not assess if the process of handwashing is compliant with the guidelines.

One potential approach is to use imaging techniques to detect fine hand movements and identify user gestures, provide feedback to the user or a central management system, with the overall goal being an automated tool that can ensure compliance with the hand washing guidelines. In advance of developing these systems, however, preliminary analysis on the hand washing process and a structured methodology was required.

The aim of this paper is to develop a classification system for WHO-hand hygiene gestures that can be deployed in a healthcare setting and assess the hand hygiene compliance in real time. Deep learning solutions were adopted for this study as they have shown promising results in other applications such as text recognition [9], sound prediction [10], and image annotation [8]. The remainder of this paper is organized as follows: Section II introduces the related work. Section III provides experiment setup; section IV explains the hand-hygiene dataset collection in detail. Section V explains ResNet architecture used for processing the video-based dataset. Section VI and VII presents the experimental results and discussion.

## II. RELATED WORK

This section discusses the past research in the field of gesture recognition with various data acquisition methods such as 3D sensors/wearable devices, and cameras. The necessity of having a robust hand hygiene data was identified at an early stage and an experiment setup was built for hand hygiene data recording. The objective behind the video data collection was to explore the paradigms of transfer learning (transfer the knowledge from one task to another) for the purpose of hand hygiene stage classification.

## A. WHO Hand Hygiene Stages

There are structured and distinct guidelines for washing hands as provided by health care authorities, WHO. These guidelines consist of eleven sequenced stages. WHO hand hygiene stages are shown in Figure 1 with stages 2-7 directly involve hand washing, and are the focus of this paper, with the other stages related to turning on water, drying hands [3]. Stages 2-7 were carefully analyzed and recorded with the help of 30 participants in this study. For the purpose of consistency, the names/ labels of the stages are renamed and are listed in Table 1.

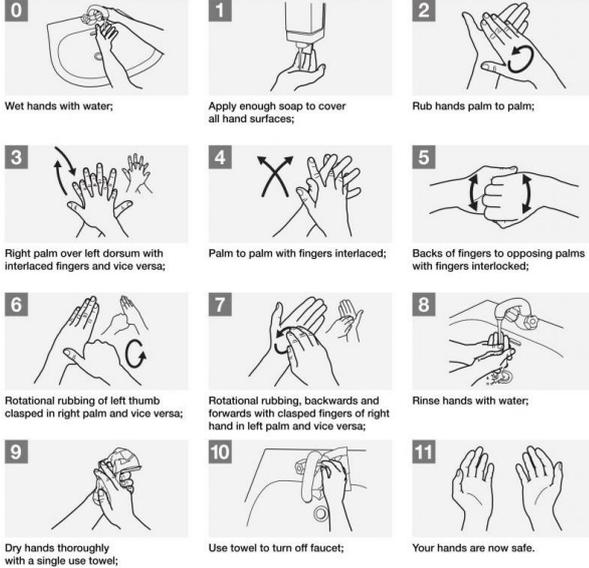

Fig 1. WHO Hand Hygiene Guidelines [3]

## B. Gesture recognition with 3D sensors

Previous work with the use of various commercial gesture tracking devices such as Leap Motion Controller, Microsoft Kinect has been carried out for the purpose of detecting hand gestures in real-time. Fabio et al. [11] extracted the hand region from the depth map and segmented it into palm and finger samples. Distance features between the palm center and the fingertips were calculated to recognize various counting gestures. Lin Shao [12] extracted the fingertip position and palm center with the help of the Leap Motion Controller for tracking hand gestures. Distance between fingertip and palm center and the distance between two fingers adjacent to each other was calculated. Velocity was detected to differentiate between static and dynamic gestures. In our previous work, hand features were identified for various hand-hygiene stages and the Leap Motion Controller was utilized to differentiate between stationary and moving hand by extracting the palm velocity vector [13]. Marin et al. [14] used the Leap Motion Controller and the Microsoft Kinect jointly to extract the hand features such as fingertip position, hand center, and hand curvature for recognizing the American Sign Language gestures. Jin et al. [15] have used multiple Leap Motion sensors to develop a hand tracking system for object manipulation where the sum of distal interphalangeal, proximal interphalangeal and metacarpophalangeal angles were taken into account. Strength of grasping and pinching were the object manipulation tasks that were recorded. 3D gesture trackers have shown promising result in the past in context of gesture recognition. However, they were not explored previously for tracking hand hygiene stages. Rashmi et al. [13] has attempted to detect basic hand hygiene stage, "Rub hands palm to palm" by extracting unique hand features such as palm orientation, palm curvature and distance between the two palms with the use of threshold values.

## C. Gesture recognition with Images

The field of computer vision and image processing is greatly explored in the context of gesture recognition in the past. Vision based systems and applications were built for gesture tracking and detection. Khan [16] used color segmentation and template matching technique to detect American Sign Language gestures. Jophin et al. [17] have developed a real time finger tracking application by identifying the red color caps on the fingers using color segmentation technique in image processing. In our previous work, color techniques based on YCBCR model were explored for hand segmentation for various hand hygiene poses [13]. Azad et al. [18] extracted the hand gesture by image segmentation and morphological operation for American Sign Language gestures. Cross –correlation coefficient was applied on the gesture to recognize it with overall 98.8 % accuracy. Chowdary et al. [19] detected the number of circles to determine the finger count in real-time, where in the scanning algorithm is independent of the size and rotation of the hand. Liorca et al. [20] has classified the hand hygiene poses using a traditional machine learning approach with a complex skin-color detection, particle-filtering model for hand tracking. In our previous work, automated hand tracking for hand hygiene stages was carried out by extracting the image contours and centroid (cx, cy) derived from the image moments [13]. The limitation of this work was the inability to classify and predict various hand hygiene stages with centroid extraction. Center of the mass was an interesting feature for real-time hand tracking. However, it provided limited information for distinguishing one stage from another hand hygiene stage. Therefore, deep learning/transfer learning technology was adapted for further processing of hand hygiene stages.

## D. Gesture recognition with Deep Learning

Deep learning is an emerging approach and has been widely applied in traditional artificial intelligence domains such as semantic parsing, transfer learning, computer vision, natural language processing and more [21]. Over the years, deep learning has gained increasing attention due to the significant low cost of computing hardware and access to high processing power (eg-GPU units) [21]. Conventional machine learning techniques were limited in their ability to process data in its natural form. For decades, constructing a machine learning system required domain expertise and fine engineering skills to design a feature extractor that can transform the raw data (example: pixel values of an image) in-to a feature vector, which is passed to a classifier for pattern recognition [22]. Deep learning models learn features directly from the data without the need for building a feature ex-tractor. Researchers have utilized deep learning/transfer learning for monitoring hand gestures. Yeung et al. [23] presented a vision-based system for hand hygiene monitoring based on deep learning. The system uses depth sensor instead of the full video data to preserve privacy. The data is classified using a convolutional neural network (CNN). Li et al. [24] conducted an investigation of CNN for gesture recognition and achieved high ac-curacy, showing that this approach is suitable for the task. Yamamoto et al. [25] use vision-based systems and a CNN for a hand-wash quality estimation. They compare the quality score of the automated system with a ground-truth data obtained using a sub-stance fluorescent under ultraviolet light. Their results show that the CNN is able to classify the washing quality with high accuracy. Ivanovs et al. [26] train the neural network on labelled hand washing dataset captured in a health care setting , apply pre trained neural network models such as MobileNetV2 and Xception with >64% accuracy in recognizing hand washing gestures. In this work, a classification model is presented for recognizing hand gestures in the hand hygiene video recordings.

*Transfer Learning- a subset of machine learning*

Deep learning models essentially require thousands of data samples and heavy computational resources such as GPU for accurate classification and prediction analysis. However, there is a branch of machine learning, popularly known as "transfer learning" that does not necessarily requires large amounts of data for evaluation. Transfer learning is a machine learning technique wherein a model developed for one task is reused for the second related task. It refers to the situation where "finding" of one setting is exploited to improve the optimization in another setting [28]. Transfer learning is usually applied to the new dataset, which is generally smaller than the original data set used to train the pre-trained model. Hussain et al. [28] applied transfer learning to train caltech face data set with 450 face images with pre-trained model on ImageNet data set. Increasing the number of training steps (epochs) increased the classification accuracy but increased the training time as well. Computational power and time were the main limitations within the study. Keras API [29] provides the most common workflow of transfer learning in context of deep learning. They are:
1. Take layers from a previously trained model
2. Freeze them to avoid destroying any of the information that they contain during the future training rounds.
3. Add some new, trainable layers on top of the frozen layers. They will learn to turn the old features into predictions on a new dataset.
4. Train the new layers on your dataset.

In this paper, these steps are the basis of constructing a model where the head of the model is replaced with a new set of fully connected layers with random initializations.

III. EXPERIMENT SETUP

An experiment setup, imitating the real world health care setting such as hospital and a nursing home was required for the onset of this project. In order to establish the setup, a science laboratory with incorporating a sink, water tap and utilities such as soap dispenser and hand towel was selected. The assessment of the process may therefore be expanded to deploy a 'hand hygiene prediction system' in a healthcare setting. The system will predict the hand hygiene movements in real time and provide feedback to the user if the hand gesture is in accordance with WHO hand hygiene guidelines.

An aluminum rig was constructed to place around the laboratory's sink in order to incorporate a camera device for recording the hand hygiene gestures. It was built with the dimensions of 1x0.8x0.8 m (LxWxH) to ensure enough space to fit on a sink and accommodate utilities such as soap dispenser and a hand towel.

The size of the rig was determined by various factors:
- The viewing range for the ELP USB camera-2.1mm lens wide angle; 60 fps
- Controlled background exposure to avoid the skin colored objects to be miss-classified as an actual skin when processing with color based models [31]. Green and white sheets were used to minimize the background information (Fig. 2, 3).
- Maintain the privacy of the user by focusing only on the hand gestures. The height of the frame was reduced from 1 m to 0.8 m in order to avoid the appearance of body organs in the frame other than the hands.
- Enough space to fit on a sink and accommodate utilities such as soap dispenser, hand-towel.
- Lightweight yet sturdy so it can be relocated if necessary.

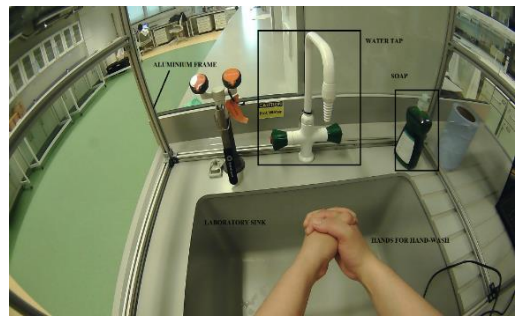

Fig 2. Hand Hygiene Data Collection Setup ( Hands in frame)

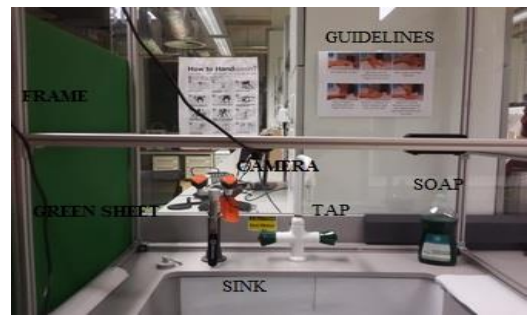

Fig 3. Background information is hidden with the help of green sheets; WHO poster is displayed so the participant does not need to memorize the hand hygiene gestures.

TABLE 1. CLASS LABELS FOR HAND HYGIENE STAGES

| Original Stage | Stage Number | Class Label |
|---|---|---|
| Rub hands Palm to Palm | 2 | Rub hands Palm to Palm (Palm2Palm) |
| Right palm over left dorsum with interlaced fingers | 3 | Fingers Interlaced |
| Palm to Palm with fingers interlaced | 4 | P2PFingersInterlaced |
| Backs of fingers with fingers interlocked | 5 | Fingers Interlocked |
| Rotational rubbing of thumb | 6 | Thumb Rub |
| Rotational rubbing with clasped fingers | 7 | Rotational Rub |

## IV. HAND HYGIENE DATASET COLLECTION

Thirty volunteers participated in this study. All the subjects were students ranging from undergraduate to postgraduate studies at Technological University, Dublin, Ireland. The required training and the demonstration about WHO hand hygiene guidelines was provided to the participants before the onset of an experiment. All participants gave their informed consent for inclusion before they participated in the study. The anonymity was ensured by capturing hand and arm movements in isolation. The video length for the hand hygiene activity was recorded for 25-30 seconds. Every hand hygiene step was followed by a pause where in the user was instructed to move their hands away from the camera. Video format for this data set is MP4 file with a size of range 40-60 MB and a frame rate of 29.84 frames/s. All of the six hand washing movements were recorded in one video for each participant. One individual frame for all the stages as a sample frame extracted from the video files is shown in Figure 4. For further processing, the video files were segmented into image files and six distinct classes were prepared. Table 2 lists the total number of image files for each class along with the label of the class corresponding to the original WHO hand hygiene stage.

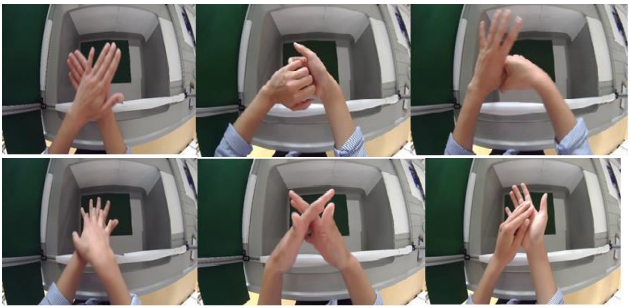

Fig. 4: Sample frames collected for hand hygiene video recordings

TABLE 2. NUMBER OF IMAGES IN EACH CLASS (320*240) SIZE

| Class Label | Number of Images |
|---|---|
| Rub hands Palm to Palm | 2,042 |
| Fingers Interlocked | 1,839 |
| Thumb Rub | 2,019 |
| P2PFingersInterlaced | 2,149 |
| Fingers Interlaced | 2,043 |
| Rotational Rub[1] | 1,834 |

[1]The data in these classes are evenly distributed in order to avoid the bias during the training of the model

## V. RESNET-50 ARCHITECTURE

ResNet architecture has gained popularity after winning the ILSVRC 2015 classification competition and COCO competitions. It is evaluated on ImageNet 2012 classification dataset that consist of 1000 classes and 1.28 million training images [32].

The ResNet architecture is based on the residual network. As the depth of the network increases, accuracy starts to become saturated and then degradation problem is exposed. The network begins to converge. Adding more layers leads to a higher training error as reported in [34]. The degradation of the network is not a result of overfitting but the initialization of the network, optimization function or gradients. The problem of degradation is addressed by a deep residual learning framework in which stacked layers fit a residual mapping instead of the original mapping. The hypothesis is that it is easier to optimize the residual mapping than to optimize the original, unreferenced mapping. Figure 5 is the building block of the residual learning where $F(x) + x$ can be realized by identity short cut connections and their output is added to the output of the stacked layers. The basis is 3X3 filters of VGGNet model as a plain network. The short cut connections are added to transform the plain network into the residual network [33].

Resnet50 is 50 layer network, a variant based on original 34 layer Resnet where each 2 layer block in 34 layer network is replaced by 3-layer bottleneck blocks to improve the accuracy and reduce the training time ; ImageNet database [33].

It consists of five convolutional layers, conv1, conv2_x, conv3_x, conv4_x, conv5_x. Once the input image is loaded, it is passed through a convolutional layer with 64 filters and a kernel size of 7*7 (conv1 layer) followed by max pooling layers. From conv2_x, the layers are grouped in pairs until the fifth convolutional layer because of the nature of the residual networks. Average pooling is applied at the fully connected layer, followed by the softmax for classification [35].

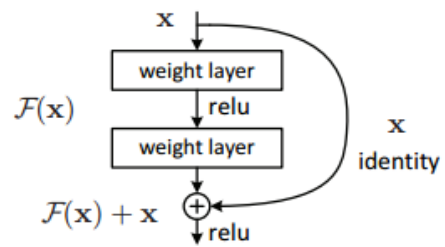

Fig 5. A building block for residual learning [33]

In this work, a pre-trained CNN model, ResNet50 [50 layers deep] on ImageNet weights is applied. The network is pre- trained on more than one million images of ImageNet dataset and can classify images into 1000 object categories. Weights="imagenet";include_top= False is selected as the head of the model was replaced with a new set of fully connected layers with random initializations. All layers below the head are frozen so that their weights cannot be updated. 'layer.trainable = False'. The model implementation is adapted from [30] where the author has implemented a multi classification system for 'sports' related video recordings and has achieved an accuracy higher than 90%.

## VI. EXPERIMENTAL RESULTS

The first set of experiment, Set 1 with images in multi classes- Fingers Interlaced, P2PFingersInterlaced and Rotational Rub were passed as an input to Resnet-50 network for 25 epochs. The training time consumed was ~15 hrs. Figure 6 is the loss-accuracy curve achieved after training the model. Cross-entropy loss is the default loss function for a multi-class classification problem. It can be specified in Keras library as 'categori-cal_crossentropy' when compiling the model. It can been seen that the network has converged with reasonable high loss for training and validation set. The accuracy <50% results in the incorrect class predictions for hand hygiene video recordings and can be seen in Figure 8.

For the second set of experiment, Set 2, the images in multi class- Rub Palm to Palm, Thumb Rub and Fingers Interlocked were passed as an input to ResNet-50 network for 50 epochs/ training steps. The training time taken was ~ 52 hrs. Figure 6, 7 is the loss-accuracy curve for training and validation set for Set 1 and 2 respectively and it can be observed that Set 2 loss is reasonable low in comparison to Set 1. Increase in the number of epochs/ training steps resulted in decrease in loss and higher accuracy>70%. The evidence can be seen in Figure 8 with correct class predictions for hand hygiene video recordings [top]. Predictions for the few selected video frames is shown in Figure 8. The complete dataset with python code and results can be accessed from the online repository that is created for this project. The link is shared in the supplementary materials.

Table 3, 4 is the classification report for Set 1 and 2 after the completion of the training process and the model is saved. Scikit-learn, an open source library for machine learning is used to compute the metrics such as precision, recall, f1 score and support. The precision is the ratio of TP / (TP + FP) where TP is the number of true positives and FP is the number of false positives. Precision is the ability of the classifier not to label a negative sample as positive. The recall is the ratio TP / (TP + FN) where TP is the number of true positives and FN is the number of false negatives. The recall is intuitively the ability of the classifier to find all the positive samples.

The F-beta score can be interpreted as a weighted harmonic mean of the precision and re-call, where an F-beta score reaches the best value at 1 and worst score at 0.
The support is the number of occurrences of each class in y_true (Target values) [36].

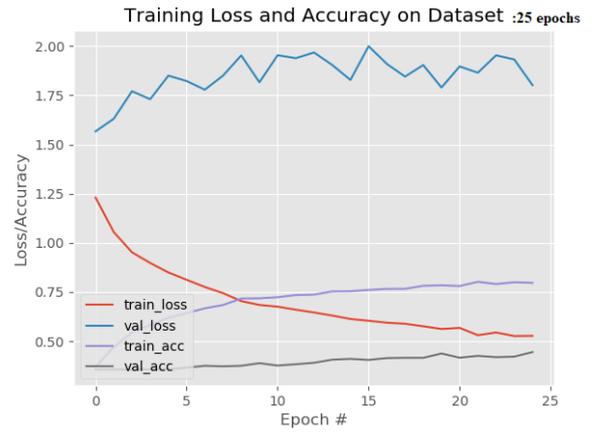

Fig 6. Accuracy/Loss curve with 25 epochs.

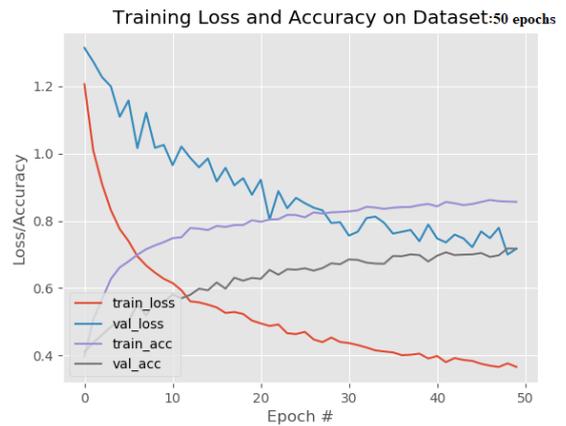

Fig 7. Accuracy/Loss curve with 50 epochs

Table 3. Classification Report for Set 1

| Class Label | Precision | Recall | F1-score | Support |
|---|---|---|---|---|
| Fingers Interlaced | 0.95 | 0.26 | 0.41 | 511 |
| P2PFingersInterlaced | 0.39 | 0.99 | 0.56 | 537 |
| Rotational Rub | 1.00 | 0.00 | 0.01 | 459 |
| Micro average | 0.44 | 0.44 | 0.44 | 1507 |
| Macro average | 0.78 | 0.42 | 0.33 | 1507 |
| Weighted average | 0.77 | **0.44** | 0.34 | 1507 |

Table 4. Classification Report for Set 2

| Class Label | Precision | Recall | F1-score | Support |
|---|---|---|---|---|
| Rub hands Palm to Palm | 0.89 | 0.88 | 0.88 | 460 |
| Fingers Interlocked | 0.91 | 0.39 | 0.54 | 510 |
| Thumb Rub | 0.57 | 0.90 | 0.70 | 505 |
| Micro average | 0.72 | 0.72 | 0.72 | 1475 |
| Macro average | 0.79 | 0.72 | 0.71 | 1475 |
| Weighted average | 0.79 | **0.72** | 0.70 | 1475 |

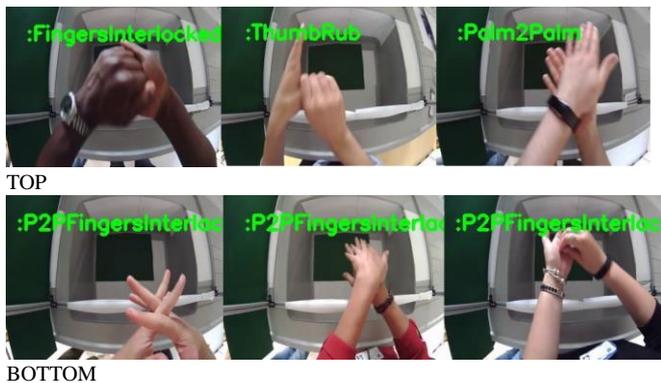

Fig 8. Predictions for sample frames. [Three at the top are correct predictions (Set 2-50 epochs). Incorrect predictions at the bottom (Set 1-25 epochs)]

## VII. Discussion

Gesture recognition and interaction has remained a popular area of research with many creative discoveries established in the past. Motion based game controllers such as Leap Motion Controller, Microsoft Kinect can track hand and arm movements thereby providing gesture interaction in real-time. However, besides the entertainment industry and other industries, such as manufacturing and automation, gesture interaction is expanding in the healthcare sector for patients with mobility and other health-related problems. One example where the tracking and identification of gestures are of interest is the process of hand washing. The process of hand washing involves dynamic hand hygiene gestures. The paper presents 'WHO-hand hygiene gesture classification system' with preliminary results based on ResNet-50 framework, applied on hand hygiene video recordings. It is noted from the results that the system performs better in class prediction with higher number of training steps. However, increase in the training steps result in a longer period of training. The training time for the data in Set 2 with 50 epochs, consumed 52 hrs. The accuracy level achieved was >70 %. The accuracy-speed trade-off is the main attribute of deploying deep learning solutions. In future, more sophisticated workstation; NVIDIA GPU with increased storage and memory will be utilized. Experiments will be conducted with larger data sets; recordings of the professional health care workers will be incorporated. ResNet-101, ResNet-152 with deeper layers will be explored to improve the performance of the system and to predict the hand hygiene stages in real-time.

*Supplementary Materials*: The complete robust hand hygiene dataset recorded for 30 participants along with the python code; model and results for the following are available online:

https://tudublin-my.sharepoint.com/:f:/r/personal/d16126930_mytudublin_ie/Documents/Hand%20Hygiene%20Research/HandWashData?csf=1&web=1&e=mMwzfp.

## Conflict of Interest

The authors declare no conflict of interest.